\documentclass{article}

\usepackage{microtype}
\usepackage{graphicx}
\usepackage{subcaption}
\usepackage{booktabs} 

\usepackage{hyperref}
\usepackage{tabularx}
\usepackage{ragged2e}
\usepackage{makecell}
\usepackage{balance} 

\usepackage[preprint]{icml2026}

\usepackage{amsmath}
\usepackage{amssymb}
\usepackage{mathtools}
\usepackage{amsthm}

\usepackage[capitalize,noabbrev]{cleveref}

\theoremstyle{plain}

\theoremstyle{definition}

\theoremstyle{remark}

\usepackage[textsize=tiny]{todonotes}

\icmltitlerunning{The $qs$ Inequality}

\begin{document}

\twocolumn[
\icmltitle{The $qs$ Inequality: Quantifying the Double Penalty of Mixture-of-Experts at Inference}

\begin{icmlauthorlist}
\icmlauthor{Vignesh Adhinarayanan}{amd}
\icmlauthor{Nuwan Jayasena}{amd}
\end{icmlauthorlist}

\icmlaffiliation{amd}{AMD Research and Advanced Development (RAD)}

\icmlcorrespondingauthor{Vignesh Adhinarayanan}{vignesh.adhinarayanan@amd.com}
\icmlcorrespondingauthor{Nuwan Jayasena}{nuwan.jayasena@amd.com}

  \icmlkeywords{Mixture-of-Experts, LLM inference, systems, reuse, memory bandwidth}

  \vskip 0.3in
]



\printAffiliationsAndNotice{}  

\begin{abstract}
Mixture-of-Experts (MoE) models deliver high quality at low training FLOPs, but this efficiency often vanishes at inference. 
We identify a \textit{double penalty} that structurally disadvantages MoE architectures during decoding: first, expert routing fragments microbatches and reduces weight reuse; second, massive resident expert pools reduce high-bandwidth memory (HBM) headroom for the KV cache. 
This phenomenon, formalized as \textit{reuse fragmentation}, pushes feed-forward networks (FFNs) into a bandwidth-bound regime, especially at long context lengths. 

We \textbf{introduce the $qs$ inequality}, a predictive criterion that identifies when MoE is structurally disadvantaged relative to a quality-matched dense model. 
This criterion unifies \textbf{sparsity ($s$)}, the fraction of parameters activated per token, and the \textbf{quality-equivalence factor ($q$)}, the size multiplier required for a dense model to match MoE performance. 
Our evaluation across frontier models including DeepSeek-V3, Qwen3-235B, Grok-1, and Switch-C demonstrates that this fragmentation is a general architectural phenomenon. 
For DeepSeek-V3 at 128k context, this results in a 4.5x throughput advantage for a quality-matched dense baseline. 
Crucially, massive architectures like Switch-C can become \textit{infeasible} on cluster sizes where a quality-matched dense model remains viable. 

Our results suggest that training-time FLOP efficiency is an incomplete proxy for inference-time performance in long-context serving. 
They also indicate that MoE may be best viewed as a training-time optimization, with distillation into dense models as a possible path toward inference-efficient deployment.
\end{abstract}

\section{Introduction}
\label{sec:introduction}

Mixture-of-Experts (MoE) architectures were designed to make large-scale language model training efficient. By activating only a small subset of parameters per token, MoE decouples model capacity from per-token computation, allowing training FLOPs to remain nearly constant while total parameter count scales to hundreds of billions or more \cite{shazeer2017sparsemoe,fedus2021switch,lepikhin2020gshard}. This sparsity is not merely a theoretical benefit; in practice, MoE models consistently achieve lower validation loss than dense models trained with the same compute budget, motivating their adoption in several frontier systems \cite{du2022glam,rajbhandari2022deepspeedmoe}.

The inference setting poses a different question. Autoregressive decoding must satisfy strict latency service-level objectives, operate at modest batch sizes, and retain key--value (KV) cache tensors for all previously generated tokens. In this regime, the performance bottleneck frequently shifts away from arithmetic throughput and toward memory bandwidth and data movement. As a result, reducing FLOPs per token does not necessarily reduce inference latency if it simultaneously increases the number of bytes transferred per token or prevents those bytes from being amortized across multiple tokens. Throughout this work, we focus on DRAM-based accelerators such as GPUs and TPUs, where HBM traffic dominates inference performance; while the qualitative arguments extend to SRAM-centric designs, the quantitative analysis is HBM-centric.

This observation exposes a potential mismatch. MoE is explicitly designed to reduce FLOPs, whereas inference performance is often governed by how effectively weights are reused across tokens. To reason about this regime, we introduce the \emph{reuse principle}: inference efficiency scales with the number of tokens that reuse each weight read, rather than with the number of FLOPs avoided. Dense feed-forward networks (FFNs) naturally satisfy this principle by amortizing weight fetches across an entire microbatch. MoE layers do not do so under realistic inference constraints. Expert routing partitions the microbatch across experts, reducing per-expert reuse and increasing weight traffic per token.

We show that this effect is structural. At inference, batch size is constrained by KV-cache residency, particularly at long context lengths. As batch size shrinks, MoE routing fragments microbatches across experts, causing per-expert reuse to approach unity, pushing FFN execution into a bandwidth-bound regime even when only a small fraction of dense FLOPs are executed. We refer to this mechanism as \emph{reuse fragmentation} and formalize it in Section~\ref{sec:reuse-principle}.

Reuse fragmentation has a second-order but practically significant consequence. Because the full expert pool must remain resident across the serving cluster, MoE consumes high-bandwidth memory (HBM) capacity that would otherwise be available for KV cache. This tightens the admissible batch size, which further shrinks per-expert subbatches and amplifies reuse fragmentation in a compounding loop. We quantify this structural disadvantage across frontier MoE systems including DeepSeek-V3, Qwen3-235B, Grok-1, and Switch-C.

Our central conclusion is that training-time FLOP efficiency is a poor proxy for inference-time economics. Under realistic serving constraints, particularly at long context, quality-matched dense models can achieve lower inference cost because they preserve reuse where MoE cannot. This conclusion follows from reuse fragmentation, which is structural rather than incidental. 

\paragraph{Contributions.}
\begin{itemize}
    \item We identify \textbf{weight reuse}, rather than FLOP count, as a key determinant of inference efficiency (Section~\ref{sec:reuse-principle}).
    
    \item We formalize \textbf{reuse fragmentation} as a structural consequence of expert routing at inference, whereby per-expert reuse scales as $R_{\text{moe}} \approx Bk/E$ (Sections~\ref{sec:reuse-principle}--\ref{sec:sparsity-reuse}).
    
    \item We derive the \textbf{$qs$ inequality} as a predictive boundary characterizing when MoE inference is structurally disadvantaged relative to a dense model of equal quality (Section~\ref{sec:sparsity-reuse}).
    
    \item We quantify capacity and throughput effects across frontier MoE models, demonstrating that resident expert pools reduce KV-cache headroom and that quality-matched dense models achieve up to a 5.3x throughput advantage (Sections~\ref{sec:results-capacity-reuse}--\ref{sec:throughput-results}).
\end{itemize}

These observations motivate reconsidering the relationship between training-time and inference-time architectures, a theme we return to in Section~\ref{sec:discussion}.

\section{Reuse as the Governing Principle for MoE Inference}
\label{sec:reuse-principle}

This section introduces the execution model for MoE at inference and presents a latency decomposition showing that, under realistic inference constraints, performance differences between dense and MoE models arise primarily from feed-forward weight reuse.

\subsection{MoE feed-forward execution at inference}

We contrast the inference-time execution of a MoE model with that of a dense model. In both cases, the attention block and KV-cache access are unchanged: each generated token reads the same KV state for attention regardless of whether the subsequent feed-forward computation is dense or expert-routed. The relevant difference at inference is therefore in how the feed-forward network (FFN) is executed.

A dense transformer executes a single FFN per layer, applying the same weight matrices to the entire microbatch of $B$ tokens. From the perspective of the hardware, this appears as one large, well-amortized GEMM per layer, with the cost of fetching FFN weights from memory amortized across the full batch.

An MoE layer replaces this execution pattern with a bank of $E$ independent FFNs (experts), each with its own weight matrices. A learned router computes token-to-expert assignments and selects $k$ experts per token. In an expert-parallel deployment, this induces a dispatch step: token embeddings are routed to the devices hosting the selected experts. Each expert then executes its FFN on the tokens it receives. The resulting expert outputs are scattered back to their original batch indices and reduced across the $k$ selected experts, producing an output tensor identical in shape to that of a dense FFN.

As a result, the original microbatch is partitioned into expert-local microbatches, and each expert executes its FFN only on the tokens routed to it. On average, each expert processes
\begin{equation}
B_{\text{expert}} = \frac{Bk}{E}
\end{equation}
tokens per step. When $E$ is large and $B$ is constrained, $B_{\text{expert}}$ can be single-digit or approach unity. This expression reflects average routing behavior across experts and captures the effective reuse governing weight amortization. Importantly, sparsity does not eliminate weight movement: each expert must still fetch its own weight matrices from memory before processing any tokens.

Because the full expert pool must remain resident in high-bandwidth memory, MoE also leaves less headroom for the KV cache than a comparable dense model. As context length grows, the admissible batch size shrinks, further reducing $B_{\text{expert}}$ and exacerbating the loss of amortization.

\subsection{Inference latency decomposition}

Given this execution model, per-token decode latency can be decomposed as
\begin{equation}
T_{\text{token}} = T_{\text{ffn}} + T_{\text{attn}} + T_{\text{comm,exposed}} . \label{eq:token-decomp}
\end{equation}
Here, $T_{\text{ffn}}$ denotes the latency of the feed-forward network, $T_{\text{attn}}$ captures the cost of attention computation including KV cache access, and $T_{\text{comm,exposed}}$ accounts for communication overhead that cannot be hidden by overlap.

Two properties of this decomposition are central. First, the attention term is largely identical for dense and MoE models: every token performs the same attention computation and must access the same number of keys and values regardless of how the FFN is structured. While attention includes arithmetic operations, its latency at inference is typically dominated by KV cache traffic. Second, while communication overhead depends on the parallelization strategy, it is usually secondary to the execution cost of the FFN. As a result, systematic differences in inference latency arise primarily through $T_{\text{ffn}}$.

\subsection{Reuse and bandwidth-dominated FFN execution}

FFN latency is governed by the maximum of compute time and memory time,
\begin{equation}
T_{\text{ffn}} = \max(T_{\text{ffn,comp}},\, T_{\text{ffn,hbm}}).
\end{equation}
In the bandwidth-sensitive regime, the memory term can be approximated as
\begin{equation}
T_{\text{ffn,hbm}} \approx \frac{W_{\text{ffn}}}{R \, \beta_{\text{hbm}}},
\end{equation}
where $W_{\text{ffn}}$ is the FFN weight size, $\beta_{\text{hbm}}$ is sustained memory bandwidth, and $R$ is the reuse factor.

For a dense FFN processing a microbatch of $B_{\text{dense}}$ tokens,
\begin{equation}
R_{\text{dense}} \approx B_{\text{dense}}.
\end{equation}
Under expert routing, reuse scales with the expert-local microbatch size. Let $B_{\text{moe}}$ denote the total microbatch size processed by an MoE layer across all experts. Then,
\begin{equation}
R_{\text{moe}} \approx \frac{B_{\text{moe}}\,k}{E}.
\end{equation}

Crucially, the admissible microbatch size at inference differs between the two cases. Because all expert weights must remain resident in high-bandwidth memory, MoE consumes more HBM capacity than a dense model of comparable quality. With less memory available for the key--value cache, the maximum feasible microbatch size satisfies
\begin{equation}
B_{\text{moe}} < B_{\text{dense}}.
\end{equation}

MoE is therefore penalized twice in the memory-dominated regime. First, reuse per expert is reduced by the routing factor $k/E$, where $k$ denotes the number of experts activated per token. Second, the baseline batch size over which weights are amortized is itself smaller. Combining these effects,
\begin{equation}
R_{\text{moe}} \ll R_{\text{dense}}
\end{equation}
under realistic inference constraints.

In this regime, MoE executes fewer FLOPs per token but amortizes FFN weight reads over far fewer tokens, increasing the number of FFN weight bytes moved per token and driving execution into a bandwidth-dominated regime. We refer to this effect as \emph{reuse fragmentation}. The reuse factor therefore provides a more accurate predictor of inference efficiency than sparsity or FLOP count alone. 
\section{Sparsity, Reuse, and Quality-Matched Inference Efficiency in MoE}
\label{sec:sparsity-reuse}

In this section, we combine reuse fragmentation with quality-matched comparisons to derive a simple criterion that predicts when MoE inference is structurally disadvantaged relative to dense models.

\subsection{Reuse under expert routing}

Mixture-of-Experts models are conventionally described in terms of sparsity. If a layer contains $E$ experts and each token selects $k$ of them, the fraction of FFN parameters activated per token is
\begin{equation}
s = \frac{k}{E}.
\end{equation}
This quantity is widely used to reason about training-time efficiency.

At inference, sparsity manifests as a reduction in the number of tokens processed by each expert. For a microbatch of $B$ tokens, the average expert-local batch size is
\begin{equation}
B_{\text{expert}} = Bs.
\end{equation}
Under the execution model in Section~\ref{sec:reuse-principle}, this implies that FFN weight reuse under expert routing scales linearly with $s$. We therefore use $s$ as a compact parameterization of reuse under MoE inference rather than as a proxy for speedup.

\subsection{Quality-matched inference comparisons}

Comparing an MoE model to a size-matched dense model is not a meaningful inference-time comparison. MoE architectures are adopted because they achieve higher accuracy under a fixed training FLOP budget. The appropriate inference baseline is therefore a dense model that matches the MoE model in output quality.

We formalize this notion using a quality-equivalence factor $q$, defined such that a dense model of equal quality has effective FFN size
\begin{equation}
W_{\text{dense}} = q\, W_{\text{active}}, \qquad
W_{\text{active}} = s\, W_{\text{moe}}.
\end{equation}

\subsection{Scaling laws and quality-equivalent multipliers}

Dense language models exhibit a predictable relationship between parameter count and validation loss. Empirical scaling laws show that loss decreases as a shallow power law with parameter count~\cite{kaplan2020scalinglaws}:

\begin{equation}
L(N) \propto N^{-\alpha_N}, \qquad \alpha_N \approx 0.076.
\end{equation}
We consider models trained to convergence (no early stopping) under a fixed evaluation setup. We neglect any additive irreducible loss term in this regime, which would only rescale the inferred multiplier without changing qualitative conclusions.

Let a dense baseline of size $N_{\text{base}}$ attain validation loss $L_{\text{base}}$, and let an MoE model attain loss $L_{\text{MoE}}$ under matched training compute. Under the scaling law above,
\begin{equation}
\frac{L_{\text{base}}}{L_{\text{MoE}}}
=
\left(\frac{N_{\text{dense,qe}}}{N_{\text{base}}}\right)^{\alpha_N},
\end{equation}
which yields
\begin{equation}
N_{\text{dense,qe}}
=
N_{\text{base}}
\left(\frac{L_{\text{base}}}{L_{\text{MoE}}}\right)^{1/\alpha_N}.
\end{equation}
The ratio $q \triangleq N_{\text{dense,qe}}/N_{\text{base}}$ defines a \emph{quality-equivalent dense multiplier}.

\subsection{A concrete anchor: DeepSpeed-MoE}

DeepSpeed-MoE provides a clear compute-matched comparison from which to derive a dense-equivalent multiplier~\cite{rajbhandari2022deepspeedmoe}. Under compute-matched training, a model with 1.3B active parameters augmented with MoE routing achieved validation loss 2.12, while a dense model with the same active size reached a loss of 2.34. Substituting these values yields
\begin{equation}
\frac{N_{\text{dense,qe}}}{N_{\text{base}}}
=
\left(\frac{2.34}{2.12}\right)^{1/0.076}
\approx 3.7.
\end{equation}

Similarly, a model with 350M active parameters with MoE achieved validation loss 2.29, while a dense model with the same active size reached a loss of 2.60. Therefore,
\begin{equation}
\frac{N_{\text{dense,qe}}}{N_{\text{base}}}
=
\left(\frac{2.60}{2.29}\right)^{1/0.076}
\approx 5.3.
\end{equation}

\subsection{The \texorpdfstring{$qs$}{qs} inequality}

We now combine sparsity and quality-equivalent scaling to obtain a simple criterion for MoE inference efficiency.

Inference cost is determined by the number of FFN weight bytes moved per token. Let $W_{\text{active}}$ denote the FFN weight footprint activated per token in an MoE layer, where $W_{\text{active}} = s W_{\text{moe}}$. A dense model of equal quality therefore has effective FFN size $W_{\text{dense,qe}} = q\,W_{\text{active}}$.

Under expert routing, FFN weights are reused only within expert-local batches. For a microbatch of $B$ tokens, each expert processes on average $B_{\text{expert}} = Bs$ tokens, so the active FFN weights are amortized over $Bs$ tokens, yielding a per-token cost proportional to $W_{\text{active}}/(Bs)$. In contrast, a quality-matched dense model amortizes its FFN weights over the full batch $B$, yielding a per-token cost proportional to $q\,W_{\text{active}}/B$.

Taking the ratio of per-token FFN weight traffic yields
\begin{equation}
\frac{\text{MoE}}{\text{dense (equal quality)}}
\;\approx\;
\frac{1}{qs}.
\end{equation}

This yields the \emph{$qs$ inequality}:
\begin{equation}
qs < 1.
\end{equation}
When $qs<1$, MoE moves more FFN weight bytes per token than a dense model of equal quality and is therefore FFN-bandwidth-disadvantaged at inference. Table~\ref{tab:qs-models} reports representative $q$, $s$, and $qs$ values inferred from prior MoE models~\cite{lepikhin2020gshard,du2022glam,fedus2021switch,zoph2023stmoe}, all of which satisfy $qs<1$.

\begin{table}[t]
\centering
\small
\caption{Representative sparsity $s$, quality-equivalence factor $q$, and implied $qs$ for modern MoE models.}
\label{tab:qs-models}
\begin{tabular}{lccccc}
\toprule
Model & $E$ & $k$ & $s=k/E$ & $q$ & $qs$ \\
\midrule
DeepSpeed-MoE & 128 & 1 & 0.0078 & $4\!-\!5$ & $0.03\text{--}0.04$  \\
GLaM & 64 & 2 & 0.031 & $4\!-\!6$ & $0.12\text{--}0.19$ \\
GShard & 128 & 2 & 0.016 & $5\!-\!6$ & $0.08\text{--}0.10$ \\
Switch-C & 2048 & 1 & 4.9 × $10^{-4}$ & $\sim 3$ & 0.0015 \\
ST-MoE & 64 & 2 & 0.031 & $\sim 3$ & 0.093 \\
\bottomrule
\end{tabular}
\end{table}

\section{Evaluation Methodology}
\label{sec:eval-method}

This section describes the decode-stage inference cost model and evaluation methodology used to produce all throughput results in the paper. The same framework is applied to multiple dense and MoE model instances under consistent system and hardware assumptions.

\subsection{Decode-stage cost model}
\label{sec:eval-cost-model}

We evaluate decode-stage inference using a per-token cost model that captures the dominant contributors to latency under autoregressive serving. Given a model configuration, a system configuration, and a parallelization strategy, per-token latency is computed as
\begin{equation}
T_{\text{token}} = \max(T_{\text{compute}},\, T_{\text{hbm}}) + T_{\text{comm,exposed}}.
\label{eq:eval-lat}
\end{equation}

This formulation is equivalent to the additive decomposition in Eq.~(\ref{eq:token-decomp}), but expressed at the level of execution bottlenecks. The compute term aggregates arithmetic work from projection, attention, and feed-forward network (FFN) execution, computed from closed-form FLOP counts normalized by dtype-specific peak throughput and parallelization factors (tensor, expert, and pipeline parallelism). The HBM term aggregates memory traffic from FFN weight reads and key--value (KV) cache access, divided by sustained HBM bandwidth. Because these components overlap in execution, per-token latency is governed by the slower of compute or memory, motivating the max formulation. Communication costs are modeled using collective-specific latency and bandwidth terms and are partitioned into overlappable and exposed components.

Weight reuse is modeled explicitly: dense FFNs amortize over the full batch, whereas MoE reuse scales as $Bk/E$. This reuse factor governs whether FFN execution is compute- or bandwidth-dominated.

\subsection{KV cache modeling and memory feasibility}
\label{sec:eval-kv}

KV cache memory is modeled explicitly because it dominates per-GPU memory consumption at long context and directly determines the maximum feasible batch size. For a given model and context length, the KV footprint per layer per GPU depends on the KV layout (full, GQA, MQA, or MLA-style compression) and on how the KV cache is sharded across GPUs.

KV-parallel strategies reduce the per-GPU KV footprint by partitioning the cache along either attention-head or sequence dimensions. As a result, each GPU retains only a fraction of the total KV state, while the remainder is distributed across the parallel group.

Memory feasibility is enforced via a per-GPU capacity constraint,
\begin{equation}
\text{budget} = (\text{HBM}_{\text{cap}} - \text{reserve} - \text{misc}) \cdot (1 - \text{safety}),
\label{eq:eval-budget}
\end{equation}
after accounting for all resident weights under the chosen partition. The remaining budget determines the maximum number of concurrent sequences per GPU, denoted $n_{\text{eff,max}}$. All throughput results are evaluated at batch size $B = n_{\text{eff,max}}$.

This capacity-first evaluation is central to our comparison: at long context, MoE models retain larger resident weight footprints, leaving less memory for KV cache and therefore admitting smaller feasible batches, which directly reduces FFN weight reuse.

\subsection{Communication and overlap modeling}
\label{sec:eval-comm}

Communication costs are modeled using ring-style formulas for all-reduce, all-gather, and all-to-all collectives, with explicit startup latency and bandwidth terms following standard collective cost models~\cite{thakur2005optimization,hoefler2007optimal,patarasuk2009bandwidth}. Collective families include tensor parallel, KV-parallel, expert parallel, and layout-transition collectives.

Collectives are mapped to either intra-node or inter-node fabrics based on group size and island capacity, defined as the maximum number of accelerators that can participate in a collective entirely within the node-local high-bandwidth fabric (e.g., an NVSwitch or equivalent domain). To estimate exposed communication, each collective family is associated with a phase-specific compute window (attention, projection, or FFN). Only communication that falls within these windows may overlap with compute or HBM traffic; any remainder is treated as exposed and contributes directly to per-token latency.

For context parallelism in pass-Q mode, a ring-tail portion of the collective is treated as non-overlappable to capture sequential propagation effects inherent to the ring schedule.

\subsection{Parallelization search and autotuning}
\label{sec:eval-autotune}

Throughput depends on interactions between KV feasibility, weight reuse, collective choice, and overlap. We therefore evaluate a search space of parallelization strategies and select the highest-throughput feasible configuration for each model.

The search space spans tensor, pipeline, expert, and data parallelism, along with KV-parallel choices (none, KVP~\cite{helix2025}, CP~\cite{yang2025context}) and their associated message counts. For each candidate partition, we compute feasibility ($n_{\text{eff,max}}$), per-token latency, tokens per second per GPU, and tokens per second at cluster scale.

The same search space, feasibility constraints, and selection criteria are applied symmetrically to dense and MoE models.

\subsection{Reference instantiation and sensitivity}
\label{sec:eval-inst}

Unless otherwise stated, all results use a reference system representative of an exemplary high-performance platform with high-bandwidth intra-node interconnect (1.8\,TB/s bidirectional aggregate). To ensure our conclusions are robust to underlying system assumptions, we evaluate sensitivity across additional hardware and parallelism dimensions to confirm that the observed throughput and latency trends generalize. The primary configuration and these explored dimensions are summarized in Table~\ref{tab:eval-summary}.

We evaluate the published architectures and MoE configurations of \emph{LLaMA}~\cite{touvron2023llama}, \emph{Grok-1}~\cite{xai2024grokos,xai2024grok1github}, \emph{Qwen3}~\cite{yang2025qwen3}, \emph{DeepSeek-V3}~\cite{deepseek2024v3}, and \emph{Switch Transformers} (Switch-C setting)~\cite{fedus2021switch}. For each model, we match the KV layout and routing parameters to their published specifications and evaluate decode-stage inference up to a context length of $L = 16$M tokens.

\begin{table}[h]
\centering
\small
\caption{Reference configuration and explored dimensions for evaluation.}
\label{tab:eval-summary}
\setlength{\tabcolsep}{4pt}
\renewcommand{\arraystretch}{1.2}
\begin{tabularx}{\columnwidth}{l >{\RaggedRight\arraybackslash}X >{\RaggedRight\arraybackslash}X}
\toprule
Category & Reference configuration & Explored / varied \\
\midrule
Model &
Grok-1, Qwen3-235B, DeepSeek-V3, Switch-C &
Quality-matched dense vs.\ MoE \\

Context &
L = 131,072 tokens &
1K to 16M tokens \\

Hardware &
180\,GB HBM3e, 8.0\,TB/s HBM bandwidth, 4.5\,PFLOPS FP8 / 2.25\,PFLOPS BF16 dense compute &
$\pm$25\% capacity and bandwidth \\

Interconnect &
High-bandwidth intra-node interconnect &
900\,GB/s (primary), 448\,GB/s (secondary) (uni-dir) \\

Parallelism &
TP=8,  EP=4, PP=1, KVP=4 &
TP$\in\{1,2,4,8\}$, EP$\in\{1,2,4,8\}$, PP$\in\{1,2,4\}$, KV$\in\{\text{none},\text{KVP},\text{CP}\}$  \\

Communication &
Ring collectives, phase-aware overlap &
Startup latency ($\alpha$), overlap fractions ($\beta$), CP ring tail \\

\bottomrule
\end{tabularx}
\end{table}
\section{Capacity and Reuse Results}
\label{sec:results-capacity-reuse}

We quantify the \emph{capacity-limited batch size} at long context and the resulting \emph{FFN weight reuse} under MoE inference.

\subsection{MoE--dense reuse comparison}
\label{sec:paired-reuse}

We compare each MoE model against a quality-matched dense baseline to isolate reuse effects at inference.
For an MoE with total parameters $X$, $E$ experts, and top-$k$ routing, the active parameter fraction is $s = k/E$.
The dense baseline is defined as
\begin{equation}
X_{\text{dense}} = q\, s\, X,
\qquad q = 5 .
\end{equation}

We set $q=5$ to reflect the empirically observed $4\!-\!6\times$ gap in active parameters required for dense models to match MoE validation loss under equal training compute, and apply it uniformly across all paired comparisons.

At inference, we compute the maximum feasible batch size under a fixed HBM budget, denoted $B_{\text{moe}}$ and $B_{\text{dense}}$, and define reuse as
\begin{equation}
R_{\text{dense}} \approx B_{\text{dense}},
\qquad
R_{\text{moe}} \approx B_{\text{moe}} \cdot s.
\end{equation}

The resulting reuse gap admits the factorization
\begin{equation}
\frac{R_{\text{dense}}}{R_{\text{moe}}}
=
\underbrace{\frac{E}{k}}_{\text{routing}}
\cdot
\underbrace{\frac{B_{\text{dense}}}{B_{\text{moe}}}}_{\text{capacity}},
\label{eq:reuse-factorization}
\end{equation}
which separates reuse loss due to expert routing from additional loss induced by reduced batch capacity.

\paragraph{Operating points.}
We report paired results at 32 GPUs (TP=8, KVP=4) and 64 GPUs (TP=8, KVP=8).
All results use FP8 weights, BF16 KV storage, and a 128k context.
Evaluation methodology and memory accounting are described in Section~\ref{sec:eval-method}.

\subsection{Routing dominates reuse loss; capacity modulates it}
\label{sec:routing-capacity-results}

Table~\ref{tab:paired-reuse} reports admissible batch sizes, implied reuse, and the routing--capacity decomposition from Equation~\eqref{eq:reuse-factorization}.

First, for fine-grained MoEs such as Qwen3-235B-A22B and DeepSeek-V3, routing dominates reuse loss.
At 64 GPUs, the routing factor $\frac{E}{k}$ alone reduces reuse by $16\times$ for Qwen3 and $32\times$ for DeepSeek-V3, establishing expert fragmentation as the primary driver of reuse collapse.

Second, capacity effects at long context are multiplicative but secondary at this scale.
For Qwen3, reduced KV headroom introduces an additional $1.03\times$ penalty, while for DeepSeek-V3 it contributes $1.13\times$.
Although smaller than routing effects, these penalties systematically compound sparsity-induced reuse loss.

Third, in extreme sparsity regimes, resident expert weights alone can render MoE inference infeasible at fixed cluster size. Switch-C-2048 cannot admit even a single sequence at 128k context on 64 GPUs, demonstrating that sufficiently fine expert partitioning can exceed memory capacity regardless of routing efficiency.

Coarse-grained MoEs operate differently. For Grok-1, the active fraction satisfies $qs > 1$ at $q=5$, granting the MoE a slight capacity advantage ($B_{\text{dense}}/B_{\text{moe}} = 0.90$). However, we expect the quality gap to narrow for coarse variants because their larger expert granularity leaves less unactivated weight space for a dense model to compensate for. Sensitivity sweeps at $q \in \{2, 3\}$ show that as the dense baseline becomes structurally smaller ($qs < 1$), it reclaims the capacity advantage ($B_{\text{dense}}/B_{\text{moe}} = 1.24$ at $q=2$). Critically, even in this best-case scenario, the total reuse gap remains near $5\times$, confirming that expert-routing fragmentation remains the fundamental bottleneck for MoE inference.

\begin{table*}[t]
\centering
\small
\setlength{\tabcolsep}{6pt} 
\caption{Reuse comparison between MoE models and quality-matched dense baselines (128k context).}
\label{tab:paired-reuse}
\begin{tabular}{lcccccccccccc}
\toprule
Model & GPUs & $q$ & $(E,k)$ & $s$ & $qs$ & $B_{\text{moe}}$ & $B_{\text{dense}}$ & $R_{\text{moe}}$ & $R_{\text{dense}}$ & $E/k$ & $\frac{B_\text{dense}}{B_\text{moe}}$ & $\frac{R_\text{dense}}{R_\text{moe}}$ \\
\midrule
Qwen3-235B & 64 & 5 & (128, 8) & 0.0625 & 0.313 & 86  & 89  & 5.38  & 89  & 16.0 & 1.03 & 16.48 \\
\midrule
DeepSeek-V3 & 32 & 5 & (256, 8) & 0.0313 & 0.156 & 100 & 130 & 3.12  & 130 & 32.0 & 1.30 & 41.60 \\
DeepSeek-V3 & 64 & 5 & (256, 8) & 0.0313 & 0.156 & 244 & 267 & 7.63  & 267 & 32.0 & 1.13 & 36.16 \\
\midrule
Grok-1      & 64 & 5 & (8, 2)   & 0.2500 & 1.250 & 127 & 114 & 31.75 & 114 & 4.0  & 0.90 & 3.60  \\
Grok-1 (Sens.) & 64 & 3 & (8, 2) & 0.2500 & 0.750 & 127 & 143 & 31.75 & 143 & 4.0  & 1.13 & 4.52  \\
Grok-1 (Sens.) & 64 & 2 & (8, 2) & 0.2500 & 0.500 & 127 & 158 & 31.75 & 158 & 4.0  & 1.24 & 4.96  \\
\midrule
Switch-C-2048 & 64 & 5 & (2048, 1)& 0.0005 & 0.002 & OOM & --- & ---   & --- & 2048 & ---  & ---   \\
\bottomrule
\end{tabular}
\end{table*}

\paragraph{Sensitivity to cluster scale.}
Capacity pressure increases as cluster size shrinks, a regime typical of latency-optimized deployments.
For DeepSeek-V3, reducing the cluster from 64 to 32 GPUs increases the capacity factor $\frac{B_\text{dense}}{B_\text{moe}}$ from $1.13\times$ to $1.30\times$ with unchanged routing sparsity, widening the reuse gap from $36\times$ to $42\times$.
Because this term multiplies the routing factor in Equation~\eqref{eq:reuse-factorization}, fine-grained MoEs incur disproportionately larger reuse loss on smaller clusters, precisely the setting where low latency is most desirable.

\section{Throughput Results}
\label{sec:throughput-results}

This section reports decode-stage throughput for the evaluated models across varying context lengths and relates the measured throughput gap to the capacity and reuse effects established earlier.
All results use FP8 weights, BF16 KV storage, and the evaluation methodology in Section~\ref{sec:eval-method}.
To highlight relative scaling and degradation, throughput is reported as a percentage relative to the MoE baseline at the 1k context length.

\subsection{Throughput versus context length}
\label{sec:throughput-vs-context}

Table~\ref{tab:throughput-vs-context} reports decode-stage throughput for DeepSeek-V3 on 64 GPUs across representative context lengths. Throughput is reported as a percentage relative to the MoE baseline at the 1k context length (100\%).

As context increases, KV cache growth reduces the admissible batch size for both variants, limiting FFN reuse and increasing the relative cost of KV traffic. At short context (1k), both models admit massive batch sizes. Yet, even in this best-case scenario for arithmetic efficiency, the dense baseline outperforms the MoE variant by over $2.1\times$ (achieving 214\% relative throughput versus the MoE's 100\%). This result demonstrates that reduced FLOP count does not dictate inference speed.

This gap peaks at a $5.3\times$ throughput advantage for the Dense-5 configuration at 16k tokens. In this regime, the difference in batch-level FFN weight reuse between the two models is at its most pronounced before extreme KV cache limits take over.

At long context (128k), throughput degrades for both variants as KV traffic dominates execution. Although the gap narrows slightly, Dense-5 still maintains a $4.5\times$ advantage, consistent with the reuse analysis in Section~\ref{sec:results-capacity-reuse}. Finally, at extreme context (4096k--16384k), the architectural differences between the models are completely neutralized. The massive KV cache footprint forces both models to collapse to single-sequence execution. Throughput drops below 1\% and converges entirely as FFN amortization vanishes for both architectures.

Across all regimes, the dense baseline outperforms the MoE variant whenever batch-level reuse is non-trivial. Autotuned partition choices modulate absolute throughput, but do not alter this ordering.

\begin{table}[ht]
\centering
\caption{Decode throughput vs. context on 64 GPUs. Throughput is reported as a percentage relative to the MoE baseline at the 1k context length.}
\label{tab:throughput-vs-context}
\setlength{\tabcolsep}{4pt} 
\small 
\begin{tabular}{lrrrrr}
\toprule
 &  &  & \multicolumn{2}{c}{\textbf{Rel. Tput (\%)}} & \\
\cmidrule(lr){4-5}
\textbf{Context} & $\mathbf{B_\text{moe}}$ & $\mathbf{B_\text{dense}}$ & \textbf{MoE} & \textbf{Dense-5} & \textbf{Speedup} \\
\midrule
1k & 31349 & 34263 & 100 & 214 & $2.1\times$ \\
16k & 1959 & 2141 & 31 & 163 & $5.3\times$ \\
32k & 979 & 1070 & 18 & 88 & $5.0\times$ \\
128k & 244 & 267 & 5 & 23 & $4.5\times$ \\
1024k & 30 & 33 & 0.7 & 2.9 & $4.2\times$ \\
4096k & 7 & 8 & 0.6 & 0.7 & $1.3\times$ \\
16384k & 1 & 1 & 0.2 & 0.2 & $1.0\times$ \\
\bottomrule
\end{tabular}
\end{table}

\subsection{Attribution: why the dense model is faster}
\label{sec:throughput-attribution}

To explain the throughput gap in Section~\ref{sec:throughput-results}, we decompose per-token decode time into HBM access, compute, and exposed communication.
Table~\ref{tab:throughput-attribution} reports this breakdown at representative short and long context lengths on 64 GPUs. All costs are reported as relative units, normalized against the compute time of the MoE model at the 1k context length.

\paragraph{Short context (1k).}
At short context, both variants admit massive batches.
While MoE achieves a $10\times$ arithmetic advantage over the dense baseline (1 vs.\ 10 relative units), this theoretical gain is entirely eclipsed by exposed communication.
Dispatching tokens to experts via an All-to-All collective scales proportionally with the massive batch size, resulting in a communication cost of 17 relative units---17 times the actual MoE compute cost.
Since the dense baseline requires zero exposed communication and remains purely compute-bound, the short-context throughput gap is entirely communication-driven.

\paragraph{Long context (128k).}
With KV cache growth constraining batch size, MoE execution becomes decisively HBM-bound: normalized HBM access costs 433 units per token versus just 72 units for the dense baseline.
This massive 361-unit HBM gap explains nearly the entire throughput difference.
Compute and communication costs are severely overshadowed, demonstrating that reduced FLOP count does not translate into performance gains in the long-context regime.

\begin{table}[t]
\centering
\small
\setlength{\tabcolsep}{3pt} 
\caption{Per-token decode-time attribution (64 GPUs). Costs are reported in relative latency units, normalized against the MoE compute time at the 1k context length.}
\label{tab:throughput-attribution}
\begin{tabular}{@{}llrrrr@{}}
\toprule
 & & \multicolumn{3}{c}{\textbf{Relative Latency}} & \\
 \cmidrule(lr){3-5}
\textbf{Context} & \textbf{Model} & \textbf{HBM} & \textbf{Comp.} & \textbf{Comm.} & \textbf{Gap Driver} \\
\midrule
1k   & MoE    & 3            & 1  & \textbf{17} & Comm. (+17) \\
     & Dense  & 1            & 10 & 0           & --- \\
\midrule
128k & MoE    & \textbf{433} & 32 & 23          & HBM (+361) \\
     & Dense  & 72           & 48 & 5           & --- \\
\bottomrule
\end{tabular}
\end{table}

\subsection{Generalization Beyond DeepSeek-V3}
\label{sec:throughput-generalization}

The DeepSeek-V3 results show that long-context inference throughput is governed by reuse and memory traffic rather than FLOP count. We now examine whether this behavior generalizes across other large MoE models. Table~\ref{tab:throughput-others-128k} reports throughput for other models at 128k context on 64 GPUs.

\paragraph{Fine-grained MoEs: dense models dominate at long context.}
For fine-grained MoEs such as Qwen3-235B, the quality-matched Dense-5 configuration substantially outperforms the MoE variant, achieving a $4.4\times$ throughput advantage despite nearly identical micro-batch sizes since $B_{\text{moe}}$ fragments per expert.
This mirrors the DeepSeek-V3 results and confirms that, at long context, routing-induced reuse fragmentation dominates inference performance.
Reduced arithmetic work per token does not translate into higher throughput once FFN execution becomes HBM-bound.

\paragraph{Coarse-grained MoEs: smaller gaps reflect weaker sparsity.}
Grok-1 operates in a coarser-grained sparsity regime and exhibits a smaller throughput gap.
At the nominal quality-matching factor, Dense-5 is $1.6\times$ faster than the MoE variant.
This reduced gap does not indicate improved MoE efficiency, but rather reflects weaker sparsity and additional capacity headroom that partially offsets routing-induced reuse loss.

\paragraph{Sensitivity to quality matching: lower \(q\) widens the dense advantage for Grok-1.}
To isolate the effect of quality matching, we evaluate Grok-1 under smaller values of the quality-matching factor \(q\), which are more appropriate for coarse-grained MoEs.
As \(q\) is reduced, the dense baseline admits larger batch sizes and its throughput advantage widens from $1.6\times$ to $2.0\times$ and $2.3\times$.
This sensitivity study shows that the apparent closeness between MoE and dense performance for Grok-1 is contingent on the choice of \(q\).
Under more realistic quality matching, dense models are likely to be even more favorable.

\paragraph{Extreme capacity regimes: MoE infeasibility strengthens the conclusion.}
Switch-C-2048 represents an extreme regime in which resident expert weights alone exceed available HBM at 128k context, rendering MoE inference infeasible on this cluster.

These results show that the DeepSeek-V3 behavior is not an isolated artifact, but representative of a broader class of modern MoEs, reinforcing that inference-time efficiency at long context is fundamentally reuse- and memory-bound.

\begin{table}[t]
\centering
\small
\caption{Throughput at 128k context (64 GPUs), including sensitivity to quality-matching factor $q$ for Grok-1. Throughput is normalized as a percentage relative to each model's respective MoE baseline.}
\label{tab:throughput-others-128k}
\begin{tabular}{@{}lcccc@{}}
\toprule
 & & & \multicolumn{2}{c}{\textbf{Rel. Tput (\%)}} \\
 \cmidrule(lr){4-5}
\textbf{Model} & $\mathbf{B_\text{moe}}$ & $\mathbf{B_\text{dense}}$ & \textbf{MoE} & \textbf{Dense} \\
\midrule
Qwen3-235B & 86 & 89 & 100 & 436 \\
\addlinespace
Grok-1 ($q{=}5$) & 127 & 114 & 100 & 158 \\
Grok-1 ($q{=}3$) & 127 & 143 & 100 & 204 \\
Grok-1 ($q{=}2$) & 127 & 158 & 100 & 228 \\
\addlinespace
Switch-C-2048 & infeasible & -- & -- & -- \\
\bottomrule
\end{tabular}
\end{table}

\section{Discussion}
\label{sec:discussion}

MoE is not universally disadvantaged at inference. In this section, we clarify when sparsity helps and when it does not, and what this implies for deploying MoE models.

\paragraph{When sparsity wins vs.\ when it does not.} MoE can be competitive when per-expert subbatches are large and execution remains compute-bound. In practice, current systems are constrained by exposed communication, as shown in Section~\ref{sec:throughput-attribution} at short context lengths, while dense models remain compute-bound. As a result, the FLOPs advantage of sparsity does not translate into lower latency. Realizing MoE’s potential benefit in this regime would require advances in the network stack, including lower-latency collectives and more effective overlap between communication and computation.

At the opposite extreme, when batch sizes shrink toward unity, MoE activates only a small subset of parameters per token. In this regime, the MoE model moves fewer weight bytes than the quality-matched dense baseline and can therefore achieve higher throughput. However, the context lengths required for this behavior to emerge are extremely large, on the order of tens of millions of tokens, and lie well beyond typical deployment settings.

\paragraph{MoE as a training-time optimization.}
While MoE can provide substantial efficiency benefits during training, those gains do not reliably carry over to inference once communication and memory traffic dominate.
This suggests a practical deployment strategy in which MoE is used primarily for training, followed by distillation into dense models for inference, combining the benefits of sparsity during training with the inference efficiency of dense execution.

\section{Conclusion}
\label{sec:conclusion}

MoE architectures improve training efficiency by reducing active FLOPs, but inference performance is governed by memory bandwidth and weight reuse rather than arithmetic throughput.
We formalized this shift via the reuse principle and showed that expert routing fragments batches, reduces reuse, and inflates feed-forward weight traffic, effects that are amplified at long context by KV cache residency.

We introduced the $qs$ inequality as a compact decision rule for predicting when MoE is structurally disadvantaged relative to a quality-matched dense model.
Across frontier MoE systems, this rule correctly anticipates the observed degradation in capacity and throughput with increasing context length.
Together, these results show that training-time FLOP efficiency is an unreliable proxy for inference efficiency under realistic serving constraints.

\section*{Trademark Notice}

AMD, the AMD Arrow logo, and combinations thereof are trademarks of Advanced Micro Devices, Inc. Other product names used in this publication are for identification purposes only and may be trademarks of their respective companies.

\balance

\bibliography{example_paper}
\bibliographystyle{icml2026}

\end{document}